\documentclass{article}

\usepackage{microtype}
\usepackage{graphicx}
\usepackage{subfigure}
\usepackage{booktabs} 

\usepackage{hyperref}

\usepackage{algorithmic}

\usepackage[accepted]{icml2024}

\usepackage{amsmath}
\usepackage{amssymb}
\usepackage{mathtools}
\usepackage{amsthm}

\usepackage[capitalize,noabbrev]{cleveref}

\theoremstyle{plain}

\theoremstyle{definition}

\theoremstyle{remark}

\usepackage[textsize=tiny]{todonotes}
\usepackage{multirow}

\def\ourtitle{Spectral Journey: How Transformers Predict the Shortest Path}

\def\token#1{$\langle #1 \rangle$}
\def\lg{$L(G)$}
\def\remap{\texttt{remap}}
\def\splength{\ell^*}
\def\sp{P^*}
\def\ours{{\it SLN}}
\def\oursfull{Spectral Line Navigation}
\icmltitlerunning{\ourtitle}

\begin{document}

\twocolumn[
\icmltitle{\ourtitle}



\icmlsetsymbol{equal}{*}

\begin{icmlauthorlist}
\icmlauthor{Andrew Cohen}{meta}
\icmlauthor{Andrey Gromov}{meta}
\icmlauthor{Kaiyu Yang}{meta}
\icmlauthor{Yuandong Tian}{meta}

\end{icmlauthorlist}

\icmlaffiliation{meta}{Meta AI (FAIR)}

\icmlcorrespondingauthor{Andrew Cohen}{andrewcohen@meta.com}

\icmlkeywords{Machine Learning, ICML}

\vskip 0.3in
]



\printAffiliationsAndNotice{}  

\begin{abstract}
Decoder-only transformers lead to a step-change in capability of large language models. However, opinions are mixed as to whether they are really planning or reasoning. A path to making progress in this direction is to study the model's behavior in a setting with carefully controlled data. Then interpret the learned representations and reverse-engineer the computation performed internally. We study decoder-only transformer language models trained from scratch to predict shortest paths on simple, connected and undirected graphs. In this setting, the representations and the dynamics learned by the model are interpretable. We present three major results: (1) Two-layer decoder-only language models can learn to predict shortest paths on simple, connected graphs containing up to $10$ nodes. (2) Models learn a graph embedding that is correlated with the spectral decomposition of the \emph{line graph}. (3) Following the insights, we discover a novel approximate path-finding algorithm \emph{\oursfull}\ (\ours{}) that finds shortest path by greedily selecting nodes in the space of spectral embedding of the line graph.
\end{abstract}

\section{Introduction}
\label{intro}
\begin{figure*}[h]
 \centering
 \includegraphics[width=0.97\textwidth]{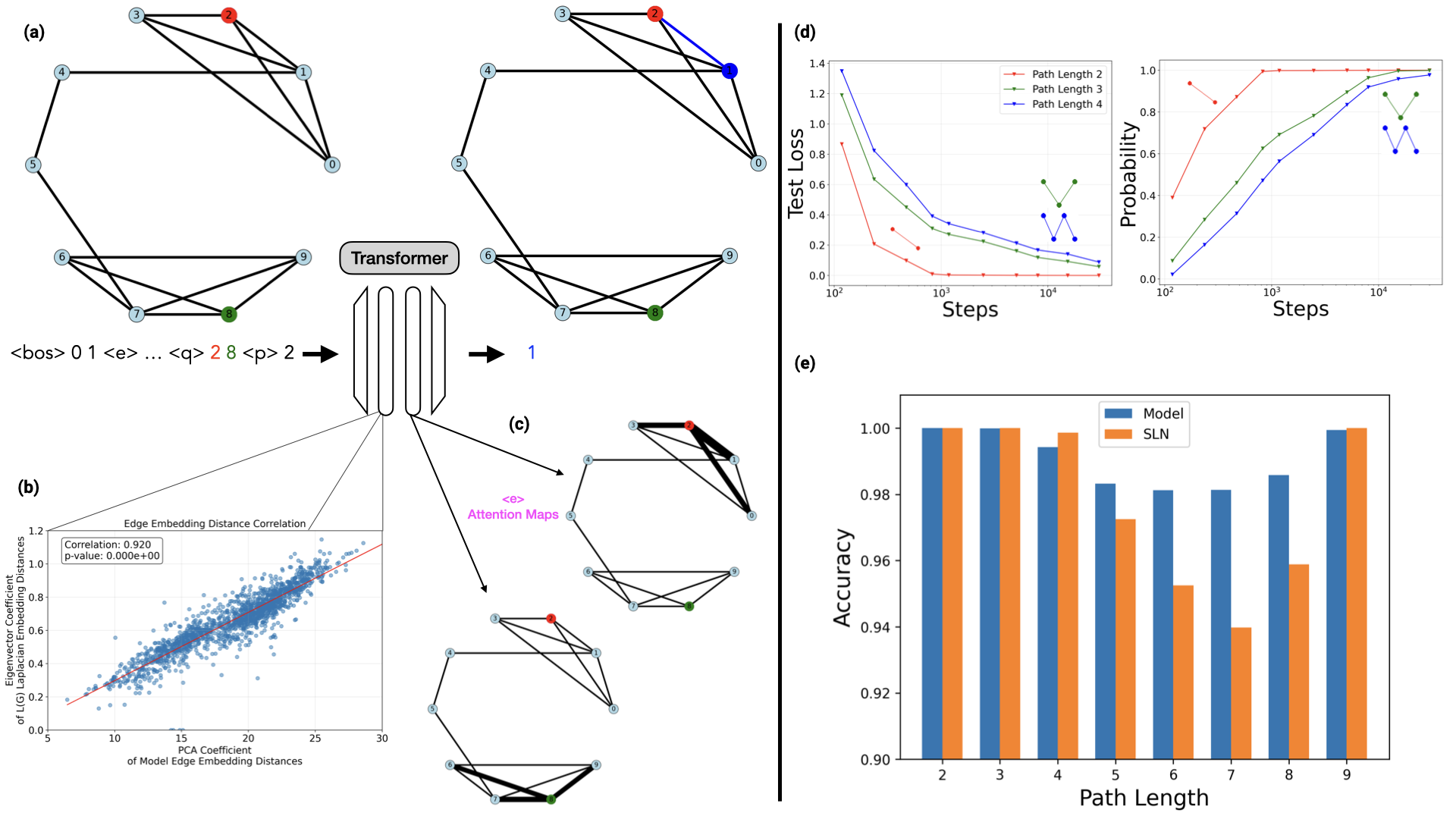}
  \caption{\small {\it Overview} {\bf (a)} We train 2-layer transformers to predict \textcolor{blue}{nodes} in the {\it shortest path} between a \textcolor{red}{source} and \textcolor{green}{target} node for a given graph represented sequentially as a list of edges and nodes, in the format of ``\texttt{<bos> 0 1 <e> 1 2 <e> ... <q> [\textcolor{red}{source}] [\textcolor{blue}{target}] <p> [\textcolor{red}{source}]}'' (i.e. ``\emph{there are edges connecting node 0 and node 1, node 1 and node 2, ..., please find shortest path between source and target}''). {\bf (b)} We find a strong correlation between model embeddings in layer 1 and the spectral decomposition of the graph and {\bf (c)} attention head dynamics (attention activations denoted by thickness of edge) in layer 2 which attend to the the current and target node edge tokens. Using this, we derive, implement, and evaluate a novel (approximate) path-finding algorithm \oursfull\ (\ours). {\bf(d)} During training, the model first learns to predict paths with 2 nodes (connected by a single edge) and then learns an algorithm for paths with $>$2 nodes. Accuracy on paths of length 3, 4 and beyond improve simultaneously. {\bf (e)} After training, the model achieves $99.42\%$ accuracy on the test set and \ours\ achieves $99.32\%$ accuracy.}\label{fig:overview}
\end{figure*}

The rise of decoder-only transformers led to a dramatic change in the capabilities and adoption of large language models~\cite{vaswani2017attention,achiam2023, claude2024,touvron2023,dubey2024}. Despite the apparent success, it remains unclear whether the models are capable of {\it planning} or {\it reasoning}~\cite{kambhampati2024, mirzadeh2024}. The lack of clarity is partially due to absence of a quantitative, verifiable definition of planning or reasoning. For this reason, language models are typically evaluated via performance on benchmarks focusing on math problems or multi-hop composition, where finding solutions requires humans to reason~\cite{cobbe2021gsm8k, yang2024twohop}. Unfortunately, these results cannot distinguish ``true reasoning/planning'' from pattern matching or retrieval due to web-scale training and possible data contamination. To make matters worse, models are often very sensitive to the exact prompt, leading to brittleness when evaluated on variations of math problems with changed variable values or irrelevant clauses~\cite{mirzadeh2024}. In other work, brittleness is shown to be correlated with test set contamination~\cite{oren2023}.

One way to progress is to go beyond benchmark numbers and study language model {\it internal dynamics} in settings with carefully controlled, structured data~\cite{YXLA2024-gsm1, edelman2024}. In such settings, it may be possible to define the notion of reasoning precisely by showing that model learns a general algorithm from a few examples. This algorithm can be extracted by interpreting the representations developed by the model. Mechanistic interpretability (MI) is a research area that aims to understand representations and reverse-engineer the computations learned by neural networks~\cite{elhage2021mathematical}.
It has revealed that models can learn general algorithms and representations in algorithmic tasks such as modular arithmetic ~\cite{power2022grokking, nanda2023,gromov2023grokking,liu2022towards,he2024learning}, path-finding in directed acyclic graphs~\cite{khona2024towards}, Markov chains~\cite{nichani2024transformers}, iterative algorithms~\cite{cabannes2024} and generating complex objects in diffusion models~\cite{okawa2024compositional}. Some MI work has proved useful for language models~\cite{templeton2024scaling} and general ML tasks~\cite{liu2022omnigrok}.

We argue that investigating language model behavior (during training \emph{and} inference) in a general abstract problem space that requires reasoning, such as graphs and graph operations, may lead to a taxonomy of algorithms and computations learned by language models for reasoning tasks. Graphs can model many reasoning and planning problems, such as Trivia, navigation, and math word problems. Additionally, the wealth of research on graph theory in computer science and mathematics provides many useful ideas for analysis.

In this work, we study GPT-style transformer models trained from scratch to predict shortest paths on simple, connected, and undirected graphs. Then, we inspect the attention maps and learned representations to identify the algorithm employed by the model to compute the shortest path. Surprisingly, we find that two-layer models can learn to perform this task on graphs with up to 10 nodes, while one-layer models cannot. Mechanistic analysis shows {\it no evidence} of traditional dynamic programming approaches~\cite{dijkstras, bellman, brinkmann2024}. Instead, we find evidence of an algorithm that is \emph{spectral} in nature~\cite{chung1997}. At a high level, the model learns an embedding scheme of edges which is correlated with their distance in the graph and then selects nodes using the minimum of this distance. We then implement this algorithm directly (without relying on the neural network) 
and evaluate it on the test set (described in the following sections) getting $99.32\%$ accuracy. We refer to this spectral algorithm as \oursfull\ (\ours). Although spectral methods for finding shortest paths have been (briefly and recently) studied in the graph theory literature~\cite{Steinerberger2020}, to the best of our knowledge, \ours\ is a novel (approximate) shortest path finding algorithm. Similar algorithmic ideas have been recently discovered in previous MI work~\cite{khona2024towards}.

The major results of this work are as follows:
\begin{itemize}
\item Two-layer decoder-only transformer models can learn to predict shortest paths on simple, connected graphs containing up to $10$ nodes. Furthermore, models with a single attention head can learn the task, but increasing the headcount while keeping the number of parameters fixed allows the model to learn the task faster.
\item Models learn a graph embedding that is correlated with the spectral decomposition of the line graph\footnote{A \emph{line graph} is a graph obtained from a given graph by replacing all edges by nodes and adding edges if the corresponding edges in the original graph share a node}. Furthermore, models strongly attend to the edges containing the current node and the target node when selecting a next edge in the path. 
\item An approximate path-finding algorithm {\ours}, which uses the distance between edges in the spectral decomposition of the line graph to compute shortest paths.
\end{itemize}

\section{Preliminaries}\label{preliminaries}
In this work, we study simple, connected and undirected graphs. We do not make any restrictions for cycles. A graph $G$ is defined as $G=(V,E)$ where $V$ is a set of nodes and $E$ is a set of edges $E=\{(v_i, v_j)\ |\ v_i,v_j \in V,\ i\ne j\}$. We denote an edge between nodes $v_i,v_j$ as $e_{i,j}$.

\paragraph{Shortest Path Problem}
Since we study connected graphs, for a source node $v_{src}$ and target node $v_{tgt}$, $v_{src} \ne v_{tgt}$, there exists a sequence of nodes or {\it path} which connects them $P=(v_{src}, v_{i},v_{i+1},...,v_{j}, v_{tgt})$ where $e_{src,i},\ e_{i,i+1},\ e_{j,tgt} \in E$. We refer to the pair $(v_{src}, v_{tgt})$ as the {\it query} and we define the length $\ell$ of path $P$ as the number of nodes in $P$. 

The shortest path problem is to find the path with the fewest nodes between $v_{src}$ and $v_{tgt}$. We define the shortest path as $\sp$ and the length of the shortest path $\splength$. Note, $\sp$ may not be unique. Computing shortest paths is a well-studied problem in graph theory and computer science and many algorithms exist~\cite{cherkassky1996}. In general, finding shortest paths is a challenging optimization problem as there may be many paths connecting two nodes. Selecting between these paths for the shortest requires non-trivial planning and reasoning.

\paragraph{Line Graph}
The Line Graph of a graph is another graph $L(G)=(V_L, E_L)$ which represents the adjacencies between edges~\cite{harary1960}. Each edge in $G$ is represented by a node in {\lg} and two nodes are connected in {\lg} if the edges share a common node in $G$. Formally, {\lg} is constructed as:
\begin{itemize}
    \item $V_L = \{v_{ij}\ |\ e_{ij} \in E\}$
    \item $E_L = \{e_{ijk}\ |\ e_{ij}, e_{jk} \in E\}$
\end{itemize}

\paragraph{Graph Laplacian and Spectral Decomposition}
The Laplacian $L$ of a graph $G=(V,E)$ is a matrix representation defined as
\begin{equation*}
L = D - A
\end{equation*}
where $D$ is the diagonal degree matrix of nodes in $G$ and $A$ is the adjacency matrix of $G$. Nodes with high degree will have a large impact on the spectrum of $L$ so it is common to consider the {\it normalized} Laplacian $\bar{L}$ defined as
\begin{align*}
\bar{L} = D^{-\frac{1}{2}} L D^{-\frac{1}{2}} \\
D^{-\frac{1}{2}}_{i,j} = \begin{cases} \frac{1}{\sqrt{deg(v_i)}} & i=j \\ 0 & i\ne j \end{cases}
\end{align*}
where $deg(v_i)$ is the degree of node $v_i$. Then, 
\begin{align*}
\bar{L}_{i,j} = \begin{cases} 1 & i=j \\ \frac{-1}{\sqrt{deg(v_i)deg(v_j)}} & i\ne j,\ e_{ij} \in E \\ 0 & else  \end{cases}
\end{align*}

The spectral decomposition of $\bar{L}$ (and $L$) and is a standard method for quantifying node connectivity for such tasks as finding node clusters. For connected undirected graphs, the eigenvalues are always positive and real-valued. The smallest eigenvalue is always zero and the eigenvector coefficient corresponding to the second smallest eigenvalue (i.e., the {\it Fiedler Vector}) corresponds to the sparsest clustering of nodes.  For more fine-grained clustering, the eigenvector coefficients for the $k$ smallest non-zero eigenvalues is used~\cite{chung1997}.

\section{Experimental Settings}\label{setting}
\paragraph{Data} We generate simple connected graphs of $3$--$10$ nodes using the Labelled Enumeration algorithm~\cite{mckay83, mckaygraphs}. which yields approximately $12$M non-isomorphic graphs in total. In this setting, the shortest path $\sp$ has between $2$ and $10$ nodes, and thus the length $2\le\splength\le10$. We partition the graphs into $80\%$ for training and $20\%$ for testing. For each pair of nodes within a graph, we compute the shortest path for the forward and reverse directions using the Python package NetworkX~\cite{nx}.

Given a graph and query (a pair of nodes), we randomly select either the forward or reverse shortest path, but not both, to prevent the model from learning symmetry shortcuts. Although there may be multiple shortest paths between two nodes, for each sample we select only one. Samples are bucketed by both path length and the number of nodes in the graph. We then sample as uniformly as possible from these buckets given that the number of paths of a particular path length decreases as the length increases (e.g., there are fewer paths of length 9 in graphs up to 10 nodes than paths of length 4). In this manner, we generate approximately $1$M training samples and $500$K test samples where the test set contains exclusively graphs not in the training set.

\begin{figure*}[h]
 \centering
 \begin{tabular}{c}
 \includegraphics[width=0.98\textwidth]{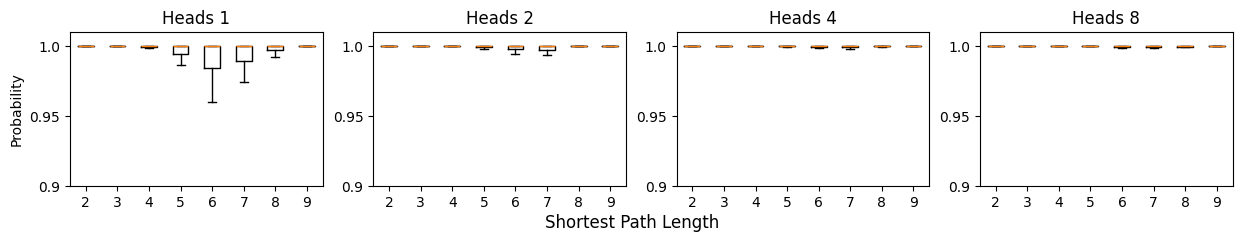}\\
 \includegraphics[width=0.98\textwidth]{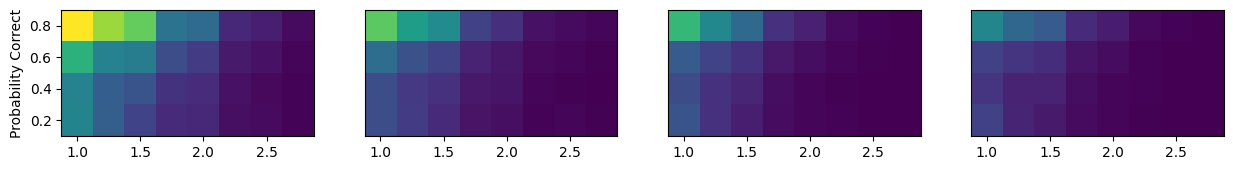}\\
 $\bar{\ell} - \splength$
 \end{tabular}
  \caption{\small ({\bf Top}) Probability of generating a shortest path by path length for 2-layer models with 1,2,4,8 heads on the test set. Increasing the number of heads improves performance, although all models are able to perform the task with high accuracy. The worst category is the 1-head model on paths of length $6$ where the 1.5 interquartile range is above $0.95$. ({\bf Bottom}) The occurrence of samples by probability of correctness and the $\bar{\ell} - \splength$ (defined in Equation~\ref{complexity}). Yellow is larger. Samples in which there are many paths of similar length between source and target ($\bar{\ell} - \splength \rightarrow 1$) contribute to the failures. As the number of heads increase, the model becomes more robust.}\label{fig:accuracy}
\end{figure*}

{\bf Representing Graphs as Tokens} Each sample is structured as a list of edges, a list of nodes, the source and target nodes and then a list of nodes corresponding to the shortest path. We use a set of $6$ control tokens to denote an edge, the beginning of the node list, the beginning of the query and the beginning of the shortest path in addition to standard beginning and end of sequence tokens. Each of the 10 node labels is a separate token thus, the vocabulary contains 16 total tokens. Please refer to Table~\ref{tab:control_toks} in Appendix~\ref{control_toks} for a complete description of the entire vocabulary.

Control tokens serve to explicitly identify the different components of a sample to relieve the model from needing to infer this from context. Additionally, control tokens serve as registers for the model to store intermediate results~\cite{goyal2024, darcet2024, brinkmann2024}.

\paragraph{Model} We train two-layer decoder-only transformer models~\cite{vaswani2017attention} with RoPE positional embeddings~\cite{RoPe}. We use a fixed hidden dimension of $512$ and train models with $1$, $2$, $4$, and $8$ heads. For a complete list of hyperparameters, please see Table~\ref{tab:hparams} in Appendix~\ref{hparams}.

During training and testing, for each sample, we randomly shuffle the edge order, the node ordering within edges and the node ordering in the node list as well as relabel each node. This maintains the same graph structure but ensures the computations and representations the model learns will be robust to any specific labeling and ordering of a particular graph. In what follows, we refer to this procedure as a \remap. Finally, we mask the loss on all but the tokens in the shortest path because the edge list, node list and source and target tokens are inherently unpredictable. 

\section{Training Results}\label{training}
In this section, we evaluate the model's ability to learn the task of generating shortest paths. As stated in Section~\ref{setting}, there may be multiple shortest paths between the source and target node. Thus, we present two results: the total probability assigned to all shortest paths and the distribution of path probabilities when many exist.  For additional experimental results ablating model graph size, please see Appendix~\ref{app:ablations}.

\begin{figure*}[h]
 \centering
 \includegraphics[width=0.98\textwidth]{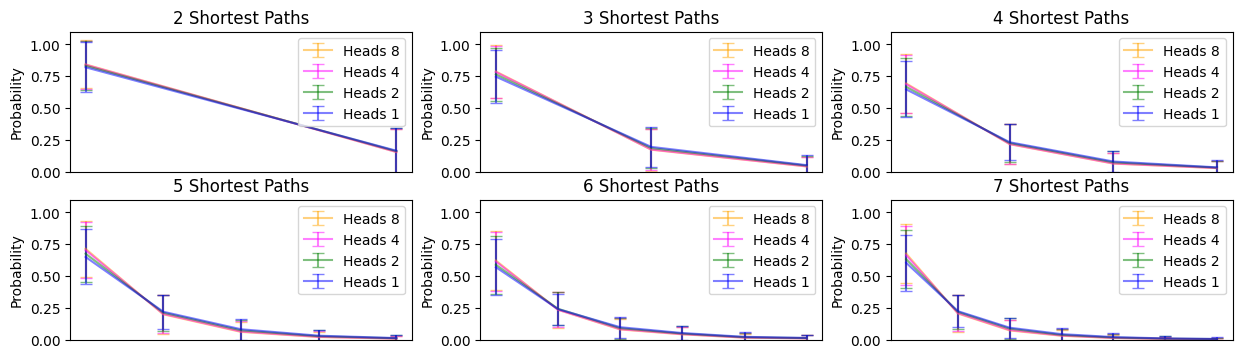}
  \caption{\small Probability distribution over $j \in \{2,3,4,5,6,7\}$ shortest paths for 2 layer models with 1,2,4, and 8 heads. Samples are grouped by the number of shortest paths between source and target in the range. We compute the probability of each of the $j$ paths and sort them in descending order. Each point is the mean and standard deviation over the test set.}\label{fig:distribution}
\end{figure*}
\subsection{Accuracy}

Figure~\ref{fig:accuracy} shows the probability of generating a shortest path resolved by the path length for 2-layer models with 1,2,4, and 8 heads on the test set. For a given graph and query, we compute all shortest paths~\cite{nx} and compute the probability of sampling each path from the model with a temperature of $\tau=0.7$. The sum of these probabilities is the probability that the model correctly generates a shortest path for a given graph and query. The distribution of probabilities over paths is presented in the next section.

In the top row of Figure~\ref{fig:accuracy}, we observe that the number of heads is positively correlated with accuracy (although all model variants perform well and the worst category is the 1-head model on paths of length $6$ where the 1.5 interquartile range is above $0.95$). Additionally, we observe that path length is not necessarily correlated with difficulty.

\begin{figure*}[h]
 \centering
 \begin{tabular}{c|c|c|c}
 $h_{current}$ & \includegraphics[width=0.25\textwidth]{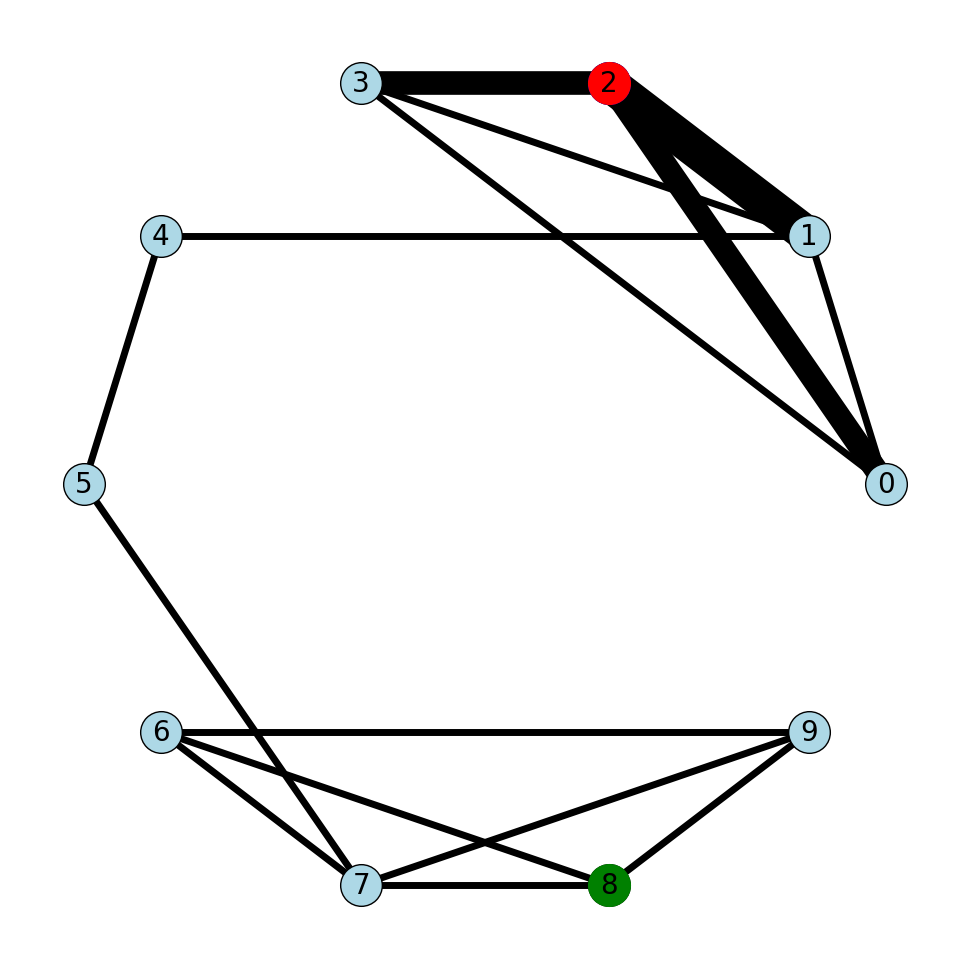} & \includegraphics[width=0.25\textwidth]{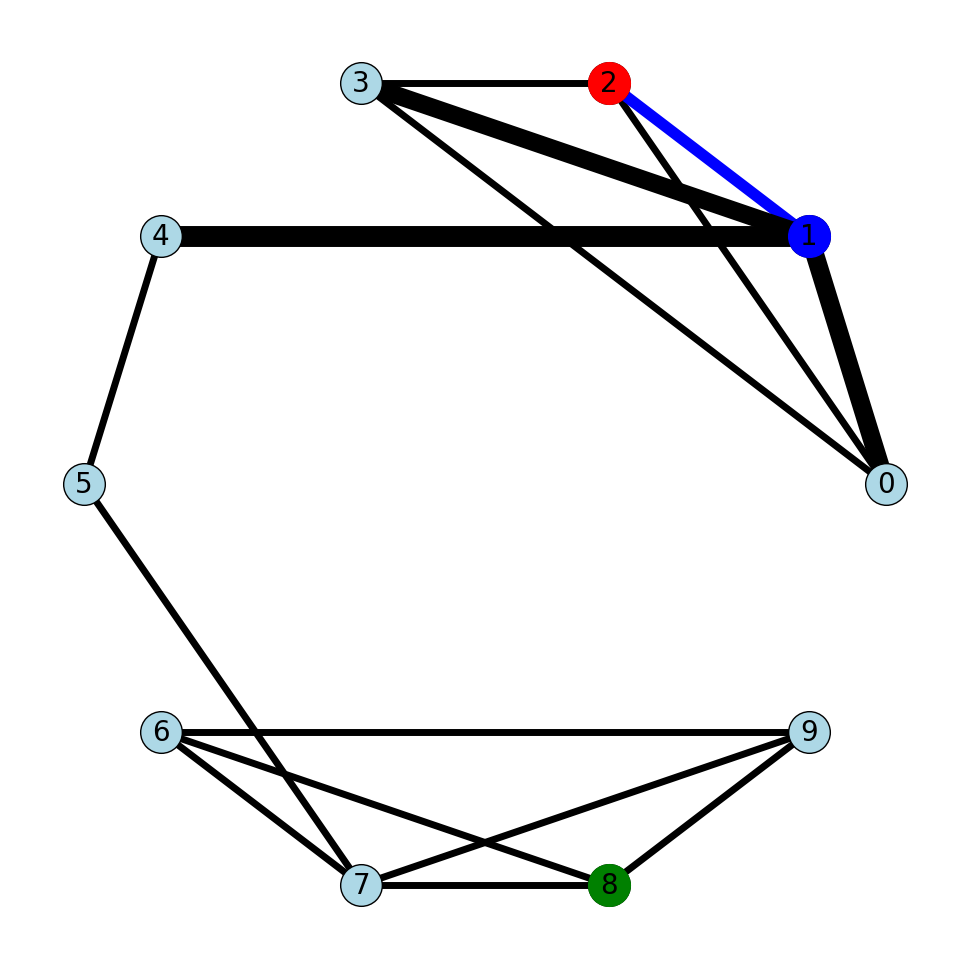} &
 \includegraphics[width=0.25\textwidth]{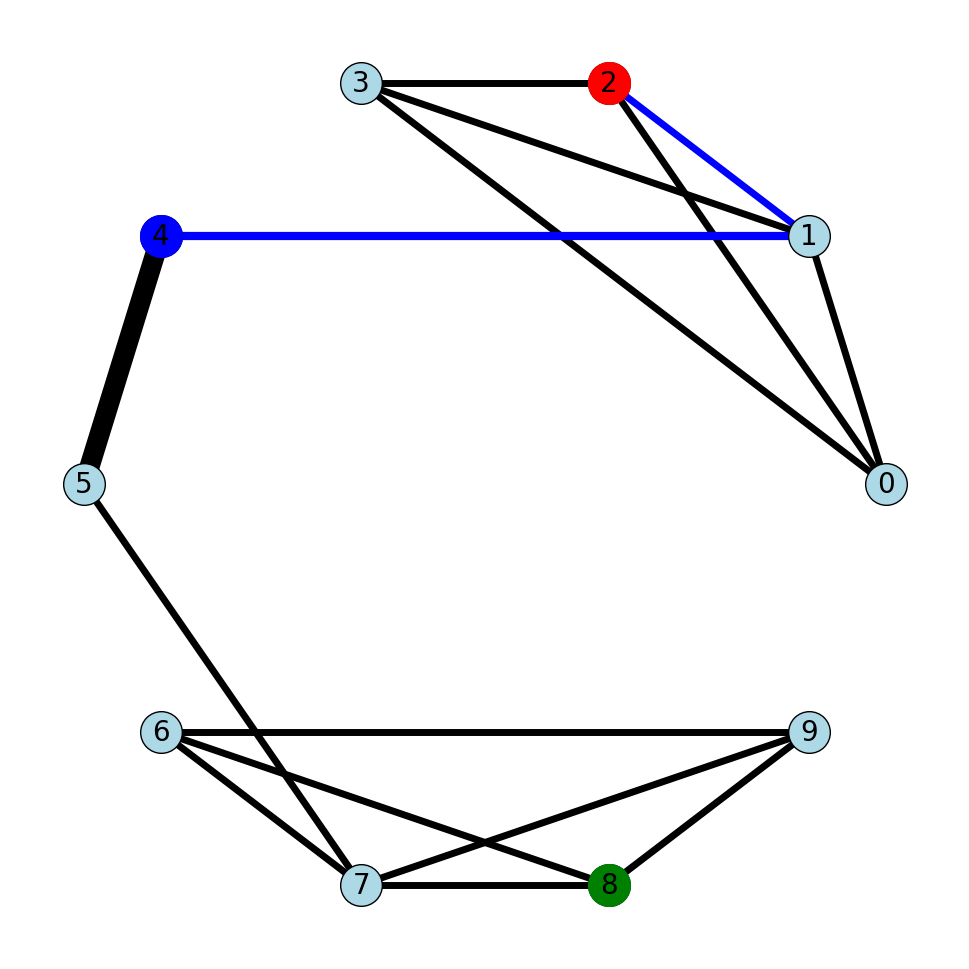} \\ \hline
 $h_{target}$ & \includegraphics[width=0.25\textwidth]{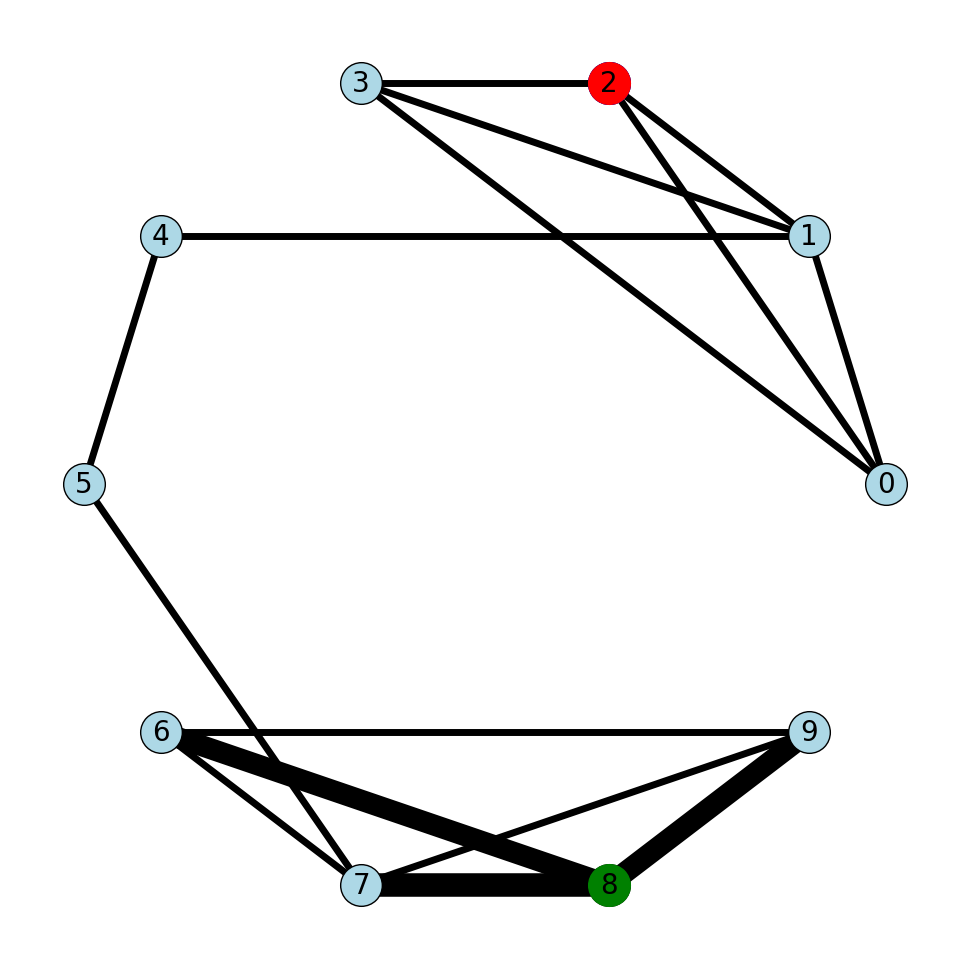} & \includegraphics[width=0.25\textwidth]{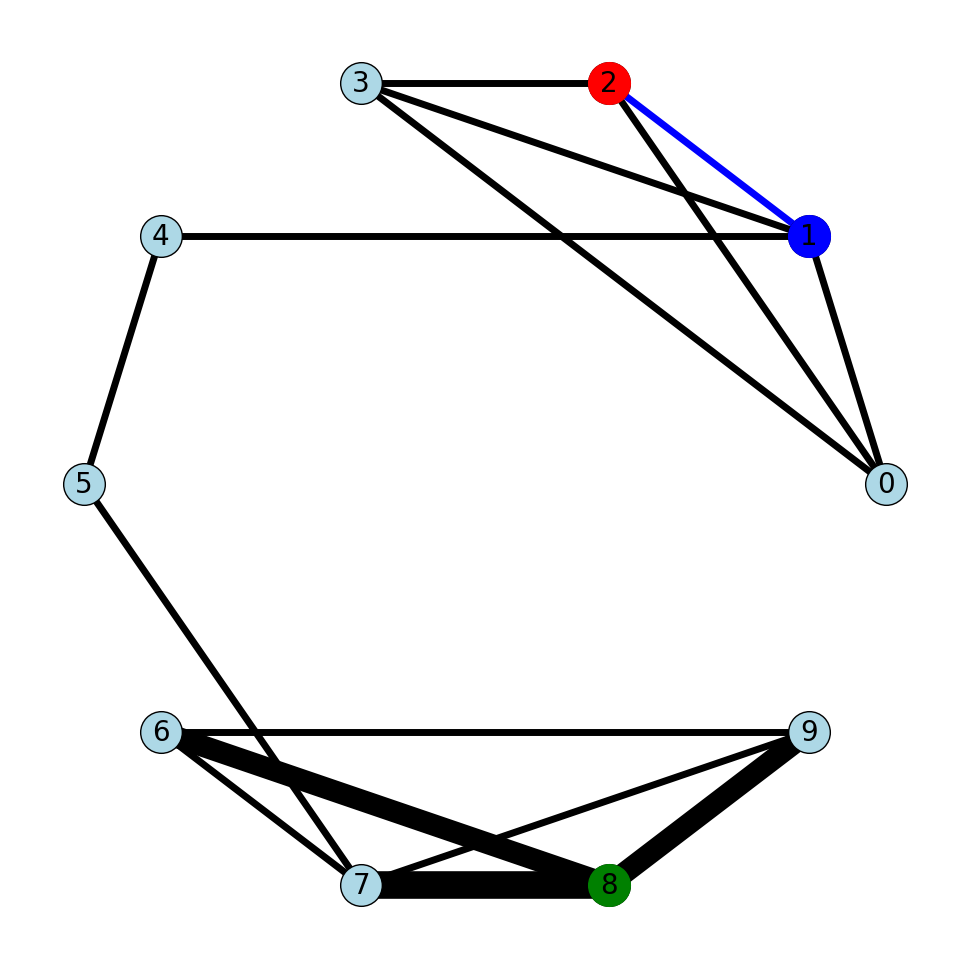} &
 \includegraphics[width=0.25\textwidth]{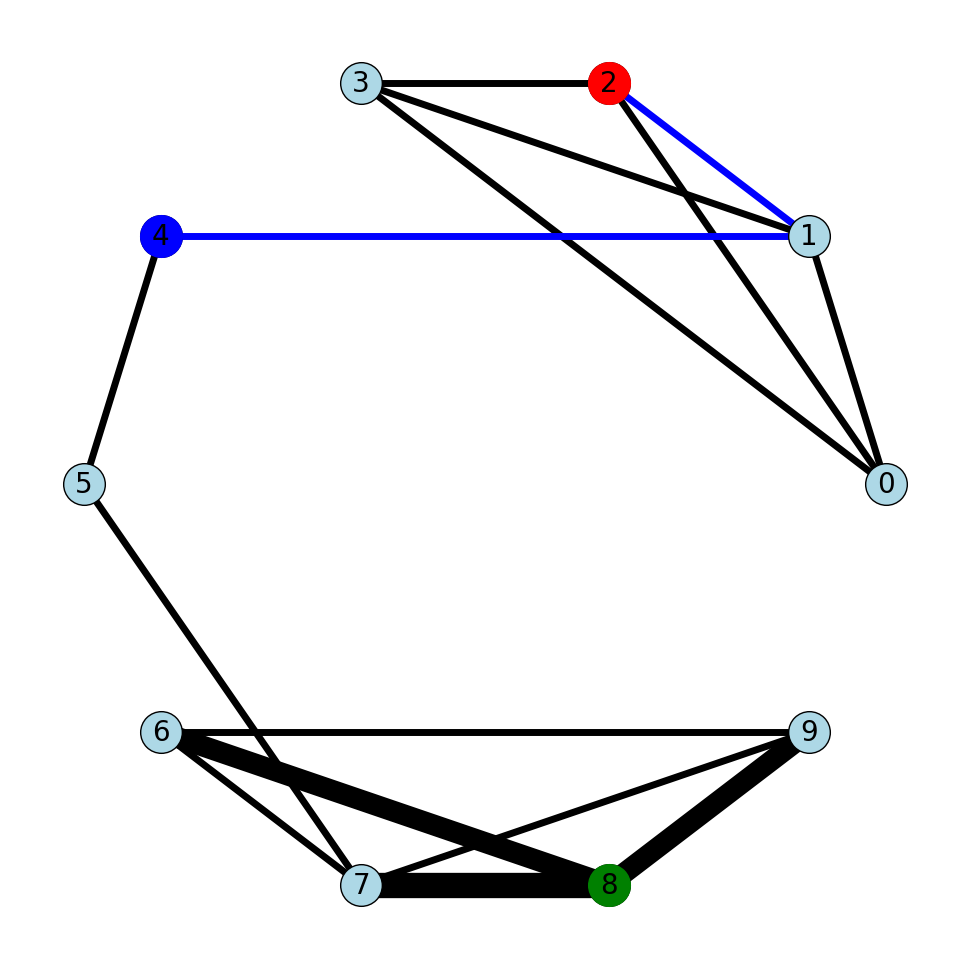} \\
 Generation & $2$ & $2$ $1$ & $2$ $1$ $4$
 \end{tabular}
 \caption{Attention activations of $h_{current}$ and $h_{target}$ of the 4 head model in the final layer visualized as thickness of an edge for an example graph and shortest path query from {\textcolor{red}{source node ($2$)}} to {\textcolor{green}{target node ($8$)}}. Each column corresponds to 1 iteration of generation and the current node and path so far is highlighted in {\textcolor{blue}{blue}}. {\bf (Top)} $h_{current}$ attends to the edge tokens corresponding to edges containing the current node in the sequence. Additionally, the relative attention activation is reduced for the edge connecting the current node to the previous node. {\bf (Bottom)} $h_{target}$ attends to the edge tokens corresponding to edges containing the target node.}\label{fig:heads_example}
\end{figure*}

In the bottom row of Figure~\ref{fig:accuracy}, we characterize the samples which have a less than $80\%$ chance of being solved by each model. In the failure cases, we observe that there usually exist many near-optimal paths (e.g., if the shortest path has length $4$, there may be other paths of length $5$) to which the model assigns non-trivial probability. For each sample, we take the $10$ paths to which the model assigns the highest probability. Then, we separate the paths with length $\ell > \splength$ and compute the average $\bar{\ell}$.

We propose the following metric to measure the variation in path length:
\begin{equation}\label{complexity}
\bar{\ell} - \splength
\end{equation}
Note, $\bar{\ell} - \splength \ge 1$ because $\bar{\ell}$ is the average path length of paths $\ell > \splength$.

We filter samples for $4 \le \splength \le 8$ as these are the samples with decreased performance. For each model, we bucket samples by the the probability assigned to the shortest paths in increments of $0.2$ up to $0.8$ and the value of $\bar{\ell} - \splength$. We plot the occurrences as color in the bottom row of Figure~\ref{fig:accuracy}.  These plots show there is a greater frequency of lower probability samples as the value of $\bar{\ell} - \splength$ approaches $1$. Additionally, models become more robust to this effect with increasing heads.

\subsection{Distribution of Paths}
In this section, we show that all model variants learn a {\it distribution} over paths even though we don't explicitly train the model on different paths for a given graph and query (up to a \remap) over epochs.  

All model variants exhibit a similar pattern - for each $k$, there is one path to which the model assigns $>50\%$ probability and then a decaying but non-trivial probability to the rest of the paths. Although this behavior emerges without explicitly training for it, it is not clear that resampling the path over training epochs would drastically change the distribution and we leave this for future work.

We group samples by the number of paths $j \in \{2,3,4,5,6,7\}$ with length $\splength$ and compute the probability for each path with each model. Then, we rank the path probabilities by descending order and average the probabilities over the rankings. We report these results in Figure~\ref{fig:distribution}.

\vspace{-0.1in}
\section{Mechanistic Investigation}\label{mechanics}
In this section, we provide two mechanistic results which we use to propose and implement {\ours} later on. Specifically, we show that the model learns attention heads which strongly attend to the edge control tokens \token{e} of edges which contain the current and target nodes in the second layer. Then, we examine the embeddings of the edge control tokens after the first layer and show that the model learns an embedding scheme wherein the distance between edge representations is correlated with the distances obtained by a spectral decomposition of {\lg}. To obtain intermediate representations from the model, we use the Python package TransformerLens~\cite{nanda2022transformerlens}.

\begin{figure*}[h]
 \centering
 \includegraphics[width=0.98\textwidth]{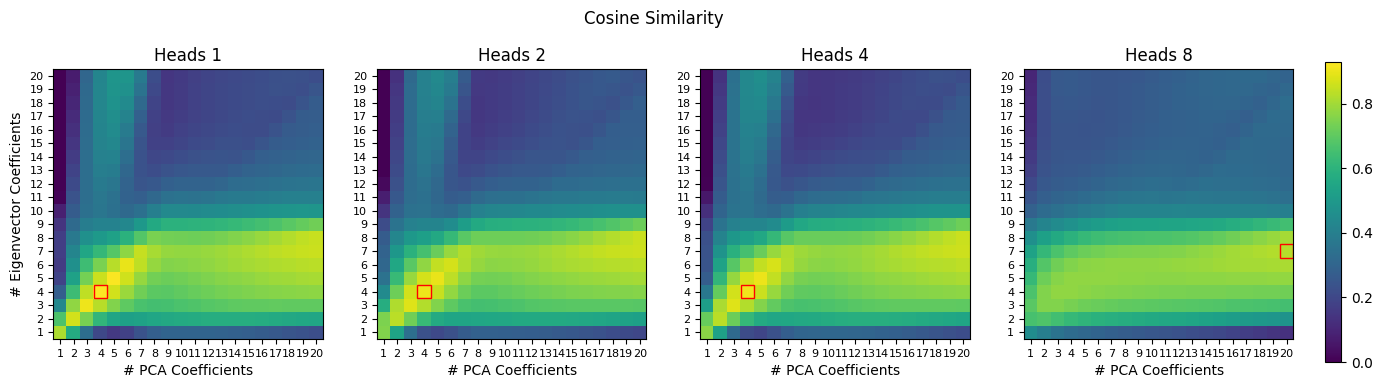}
  \caption{\small Cosine similarity of distance matrices between the top $20$ principal components of edge token embeddings and the eigenvector coefficients of the $20$ smallest non-zero eigenvalues of the normalized Laplacian of {\lg} over $10000$ random samples. For each sample, we apply \remap\ $100$ times. For 1, 2 and 4 heads, the maximums (\textcolor{red}{red square}) are $0.928$, $0.907$ and $0.909$, respectively with $4$ eigenvector and $4$ PCA coefficients. For 8 heads, the maximum is $0.826$ with $7$ eigenvector and $20$ PCA coefficients.}\label{fig:cos_sim}
\end{figure*}

\subsection{Attention Maps}\label{activations}
In Section~\ref{training}, we showed that increasing the number of heads correlated with improved accuracy. In this section, we examine attention activations as the model autoregressively generates a path between source and target node. We refer to the most recent intermediate node in the generated path as the {\it current} node.

We identify two distinct attention head mechanics in the second (and final) layer and denote them by $h_{current}$ and $h_{target}$ and define their functions as:
\begin{itemize}
\item $h_{current}$: Attends to the edge control token \token{e} of edges which contain the {\it current} node in the shortest path.
\item $h_{target}$: Attends to the edge control token \token{e} of edges which contain the {\it target} node in the shortest path.
\end{itemize}

To find $h_{current}$ and $h_{target}$, we manually inspect activation maps when the model is generating paths of at least $4$ nodes. We focus on these paths, because the attention activations for paths with $< 4$ nodes are less peaked for $h_{current}$ and $h_{target}$, suggesting the model may use a potentially simpler algorithm in these cases (e.g., a `lookup' table when the source and target nodes are connected by an edge). We select the heads from each model whose maximum activations correspond to \token{e} containing the current or target nodes. Note that the 2-head model does not need any selection, whereas the 8-head model learns redundant $h_{current}$ and $h_{target}$. Since the model distributes activation according to the degree of a node (which varies across samples), we normalize attention activations by dividing by the maximum value in a given sample. Additionally, other heads in the 4 and 8 head models attend to other control tokens as well as other edges in the graph, but we leave interpreting these activations to future work. 

In Figure~\ref{fig:heads_example}, we show the activations of $h_{current}$ and $h_{target}$ (visualized as the thickness of an edge) for an example graph as the 4-head model generates the first 3 nodes in the shortest path. In the top row, $h_{current}$ attends to the edge tokens corresponding to edges containing the current node in the sequence. Additionally, the relative attention activation is reduced for the edge connecting the current node to the previous node, which we interpret as a mechanism to avoid returning to the previous node as this would not be the shortest path. In the bottom row, $h_{target}$ attends to the edge tokens corresponding to edges containing the target node. We report the average normalized activations of $h_{current}$ and $h_{target}$ on \token{e} versus other edges, respectively {\it at the first (source) node} for paths of length 4 or greater over the test set in Table~\ref{fig:head_values} in Appendix~\ref{app:attention}.

\vspace{-0.1in}
\subsection{Spectral Partitioning of {\lg} and Model Embeddings}\label{lg_laplacian}
\vspace{-0.1in}

\begin{figure}[h]
 \centering
 \includegraphics[width=0.48\textwidth]{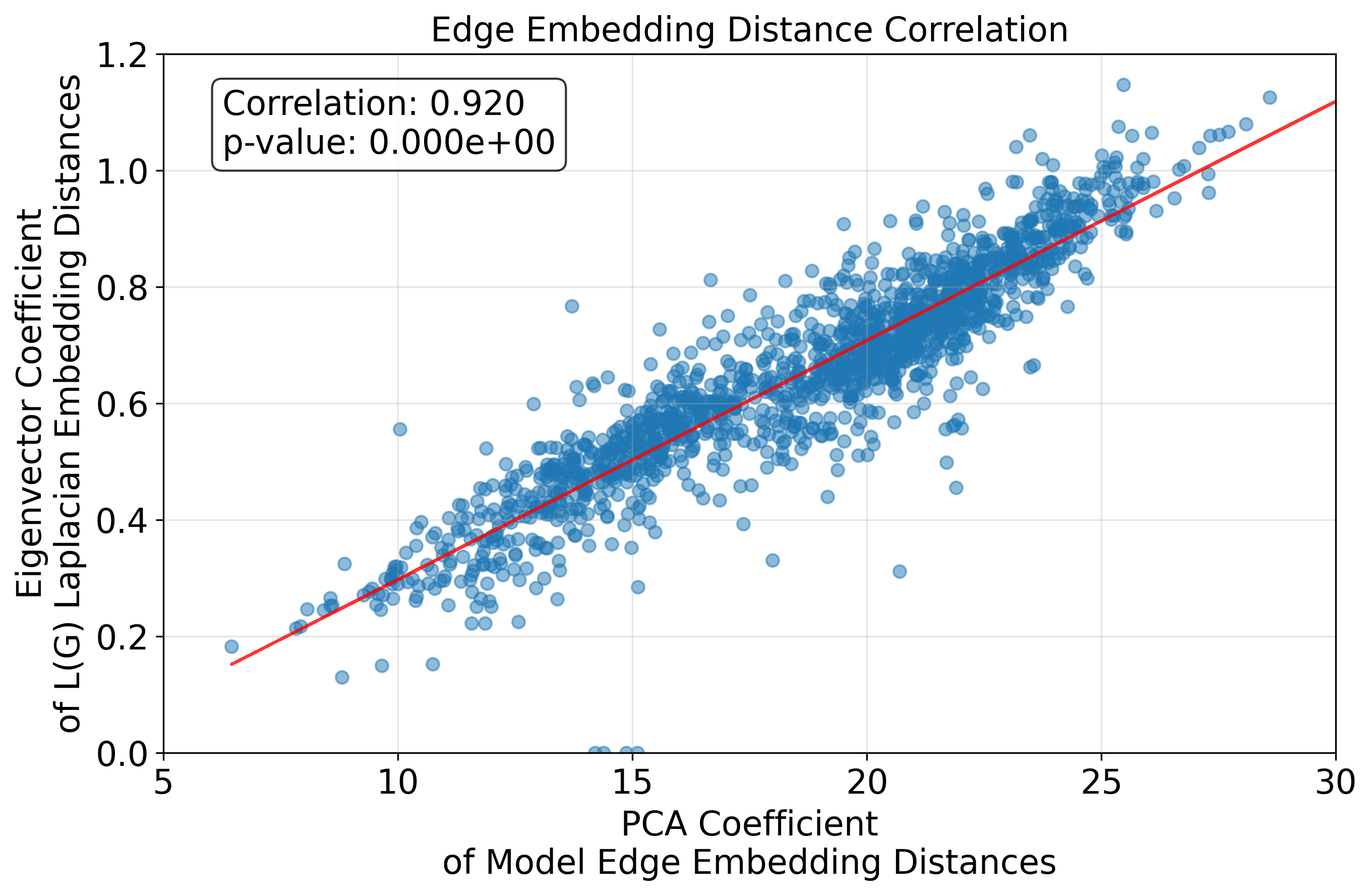}
 \vspace{-.5cm}
  \caption{({\bf x-axis}) Unnormalized pairwise distance matrix of top $4$ PCA coefficients of the embeddings of the control tokens \token{e} after the first layer of the $4$-head model versus ({\bf y-axis}) unnormalized pairwise distances between the eigenvector coefficients of edges corresponding to the smallest $4$ non-zero eigenvalues of the normalized Laplacian of {\lg}. Values are computed over all pairs of edges from 100 random samples and then we subsample $10\%$ of values. The Pearson correlation coefficient is $0.92$.}\label{fig:pairwise}
\end{figure}

Given the results in the previous section, we investigate the embeddings of the edge control token \token{e} after the first layer (as these are the representations to which the attention heads in the second layer attend). We assume that these embeddings represent specific edges (namely, the edge that precedes the control token). To interpret the algorithm learned by the model we ask: is there an algorithmic way to embed edges that comes directly from the graph theory. The most obvious guess is to consider spectral embeddings of the line graph Laplacian. Below we show that representations found by the model correlate strongly with the line graph Laplacian. Specifically, we show that the principal components of the edge embeddings are correlated to the edge representations in the spectral decomposition of {\lg}.

For a given sample, we apply \remap\ $100$ times and collect the activations of the residual stream after the first layer of the representation of \token{e}, retaining the original edge orderings to do pairwise comparisons. For each \remap, we compute the principal components separately to obtain a low-dimensional representation for each edge. We then average the pairwise distances between these representations over each \remap. Alternatively, for the original graph, we compute the spectral decomposition of the normalized Laplacian of {\lg} and use the eigenvector coefficients to obtain a representation for each edge as is commonly done in spectral graph theory. These representations are completely independent of the model. This yields two vectors of length ${n \choose 2}$ where $n$ is the number of edges in the graph. We normalize both of these vectors to the range $[-1, 1]$ and compute their cosine similarity. Note cosine similarity is in the range $[-1, 1]$ where two vectors are more correlated as the value approaches $1$.

We perform the above procedure for $10000$ random samples from the test set and compute the cosine similarity when varying the number of eigenvector coefficients and PCA coefficients. We report the results in Figure~\ref{fig:cos_sim}. For 1, 2 and 4 heads, the maximums are $0.928$, $0.907$ and $0.909$, respectively with $4$ eigenvector and $4$ PCA coefficients. For 8 heads, the maximum is $0.826$ with $7$ eigenvector and $20$ PCA coefficients.

Figure~\ref{fig:pairwise} shows the Pearson correlation between the ({\it unnormalized}) pairwise embedding distances of the top $4$ principal components of the $4$-head model's edge token representations and the eigenvector coefficients of edges corresponding to the smallest $4$ non-zero eigenvalues. These values are selected as they correspond to the maximum cell in Figure~\ref{fig:cos_sim}. We compute the pairwise distances over $100$ randomly sampled graphs and then subsample $10\%$ of points to plot. The Pearson correlation coefficient is $0.92$. In simple terms, Figure~\ref{fig:pairwise} means that (on average) if two edges are close in model-embedding space then they are also close spectra-embedding space.

We note that the \token{e} embeddings are fundamentally incapable of perfectly matching the {\lg} decomposition due to the causal mask - the specific order of the edges may make it impossible for the model to represent the precise relative distances as they may become 'closer' due to a future edge to which the model cannot attend. Averaging over \remap\ alleviates this issue. It is possible that an encoder-decoder architecture~\cite{vaswani2017attention} would be more successful but this is beyond the scope of this work as we want to understand the dynamics of decoder-only models. 

\vspace{-0.1in}
\section{Proposed Path-Finding Algorithm}
In this Section we argue that the algorithm used by the model can be interpreted completely. Since we know that model essentially learns spectral embeddings, it is reasonable to ask: can we design a simple algorithm that leverages these embeddings (and possibly distances in the embedding space) to find the shortest path. We use the insights from Section~\ref{mechanics} to propose a (novel) path-finding algorithm which we call \oursfull\ (\ours). There exist shortest path algorithms which use the graph Laplacian~\cite{Steinerberger2020}. To the best of our knowledge, \ours\ is the first which uses the Laplacian of {\lg}.

There are three key components which follow from the results in Sections~\ref{activations} and~\ref{lg_laplacian}, respectively. Given a graph $G=(V,E)$, and source and target nodes, we first compute edge embeddings of the graph edges using the spectral decomposition of the Laplacian of {\lg}. Then, beginning at the source node, we iteratively apply the following three steps until we reach the target node:
\begin{enumerate}
\item Gather the embeddings of edges containing the current node (i.e., $h_{current}$) and target node (i.e., $h_{target}$), constructing the sets $E_{current} = \{e_{current,i}\ |\ e_{current,i} \in E\}$ and $E_{target} = \{e_{target,j}\ |\ e_{target,j} \in E\}$.
\item Compute the L2 distance matrix $D^{current, target}$ between elements in $E_{current}$ and $E_{target}$
\begin{align*}
\begin{split}
&D^{current, target}_{i,j} = \lVert e_{current,i} - e_{target,j}\rVert_2\\
&e_{current,i} \in E_{current}\\
&e_{target,j} \in E_{target}
\end{split}
\end{align*}
The $i,j$'th entry $D^{current, target}_{i,j}$ is the estimate of the distance in the graph between edges $e_{current,i}$ and $e_{target,j}$.
\item Select the edge $e_{current,i}$ with minimum distance to any of the the edges in $E_{target}$.
\begin{align*}
\hat{i}\leftarrow argmin_{i,j} D^{current, target}
\end{align*}
Then, node $\hat{i}$ is the next node in the path.
\end{enumerate}


We implement and run \ours\ on the test set and are able to achieve a final accuracy of $99.32\%$. We found that for most graphs (roughly $80\%$) using only the second smallest eigenvalue (i.e., the Fiedler vector from Section~\ref{preliminaries}) for edge embeddings is sufficient. However, for other graphs we needed to increase the number $k$ of non-zero eigenvalues we consider for edge embeddings. We report the log of counts for each $1 \le k \le |E|$ in Figure~\ref{fig:k_counts} in Appendix~\ref{app:sln}. Additionally, please see Appendix~\ref{app:ablations} wherein we predict and demonstrate with ablations that the model will fail to learn \ours\ as a function of the hidden dimension and maximum number of edges in the graphs.

\vspace{-0.1in}
\section{Related Work}
\vspace{-0.05in}
{\bf Interpretability and Reasoning}
Ideas from Mechanistic Interpretability have been applied to models trained on algorithmic data to understand the types of algorithms models may discover. Solutions learned by transformers trained on modular arithmetic have been fully reverse-engineered~\cite{nanda2023} and the dynamics of feedforward networks have been theoretically characterized~\cite{gromov2023grokking, tian2024, he2024learning}. In transformers trained on data generated by iterative algorithms, the model learns a corresponding attention head which applies this iterative scheme~\cite{cabannes2024}. Circuits in transformers exist which can compose atomic subject-object relations into multi-hop relations across entities \cite{wang2024grokked} as well as compose variable relationships and assignments in propositional logic~\cite{Hong2024}.


{\bf Graph Problems with Language Models}
Graph problems have served as a useful testbed for understanding the capabilities of language models with benchmarks such as GraphQA~\cite{fatemi2024talk}, CLRS-text~\cite{deepmind2024clrstext} and NLGraph~\cite{wang2023can}. Graph problems also have been used to illuminate numerous token representation limitations for solving combinatorial problems with language models~\cite{ying2021, perozzi2024, deepmind2024clrstext, bachmann2024}.

{\bf Path-finding with Language Models} 
Interpreting language models trained on path finding tasks is of great interest as it is a problem that fundamentally requires planning and reasoning~\cite{wang2024}. Models trained on directed acyclic graphs~\cite{khona2024towards} to find any path, not necessarily the shortest, learn an algorithm that is very similar to \ours\ however it relies on node embedding distances instead of edge embedding distances, possibly due to differences in graph representation. There is no connection to the Graph or Line Graph Laplacian and we do not have any constraints on cycles in our setting.
6-layer decoder-only models trained to predict the path between source and target nodes in binary trees learn an iterative mechanism applied per layer whereas our model learns a embedding distance-based mechanism~\cite{brinkmann2024}. Exploring the dynamic between depth and algorithm one can interpret is a very interesting direction for future work.
The SearchFormer line of work shows that encoder-decoder transformers can be trained to generate $A^*$ search traces in order to navigate mazes and often generate smaller search trees than $A^*$ itself~\cite{lehnert2024beyondastar, su2024dualformer}. 
Finally, there exist many alternate lines of work such as Graph Neural Networks~\cite{wu2021gnn} and Looped Transformers~\cite{deluca2024}.

\vspace{-0.1in}
\section{Conclusion}
\vspace{-0.05in}
In this work, we have investigated the problem of training to and understanding how decoder-only language models predict shortest paths. To that end, we have shown 2-layer models can learn this task with high accuracy and, by observing attention head dynamics and demonstrating a high correlation between token embeddings and the eigendecomposition of the normalized Laplacian of the line graph, proposed a novel path-finding algorithm \oursfull. There are many possible directions for future work such as graph tasks beyond the shortest path, exploring the trade-off between model and graph size and alternative forms generalization besides unseen graph structures.


\bibliography{graphlm}
\bibliographystyle{icml2024}

\newpage
\appendix
\section{Training Details}
\subsection{Hyperameters}\label{hparams}
\begin{table}[h]
\centering
\begin{tabular}{ll}
\toprule
Field & Value \\
\midrule
layers & 2 \\
num\_heads & $\{1,2,4,8\}$ \\
hidden\_dim & $512$ \\
MLP hidden\_dim & $2048$ \\
vocab\_size & $16$ \\
RoPE $\theta$ & $10000.0$ \\
weight\_decay & $0.001$ \\
optimizer & adamw \\
$\beta_1$ & $0.9$ \\
$\beta_2$ & $0.99$ \\ 
lr\_scheduler & cosine annealing \\
lr\_ratio & .1 \\
warmup\_epochs & 1 \\
epochs & 2000 \\
\bottomrule
\end{tabular}
\caption{Training Parameters}\label{tab:hparams}
\end{table}
\subsection{Control Tokens}\label{control_toks}
\begin{table}[h]
\centering
\begin{tabular}{lp{0.8\linewidth}}
\toprule
Token & Description \\
\midrule
$[0-9]$ & Nodes \\
\token{bos} & Beginning of sequence \\
\token{eos} & End of sequence \\
\token{e} & An edge between the previous two node tokens \\
\token{n} & The beginning of the node list and end of the edge list \\
\token{q} & Beginning of the query. The following two node tokens are the {\it source} and {\it target} nodes \\
\token{p} & Beginning of the path. The following nodes are the nodes in the shortest path beginning with the {\it source} and ending with the {\it target}\\
\bottomrule
\end{tabular}
\caption{{\bf Vocabulary}}\label{tab:control_toks}
\end{table}

\section{Additional Results}
\subsection{Ablations}\label{app:ablations}
In this section, we present two additional results. Given the algorithm \ours\ which we have identified, we are able to make a prediction about the model size required to learn the task of predicting shortest paths. Specifically, since the Laplacian of the line graph is an $NxN$ matrix where $N$ is the number of {\it edges} in the graph, to fully represent this matrix, the hidden dimension of the model must be greater than $N$. In our setting, graphs contain up to $10$ nodes, so the minimum hidden dimension we would predict to learn the task is ${10 \choose 2} = 45$.
\begin{figure}[h]
 \centering
\includegraphics[width=0.45\textwidth]{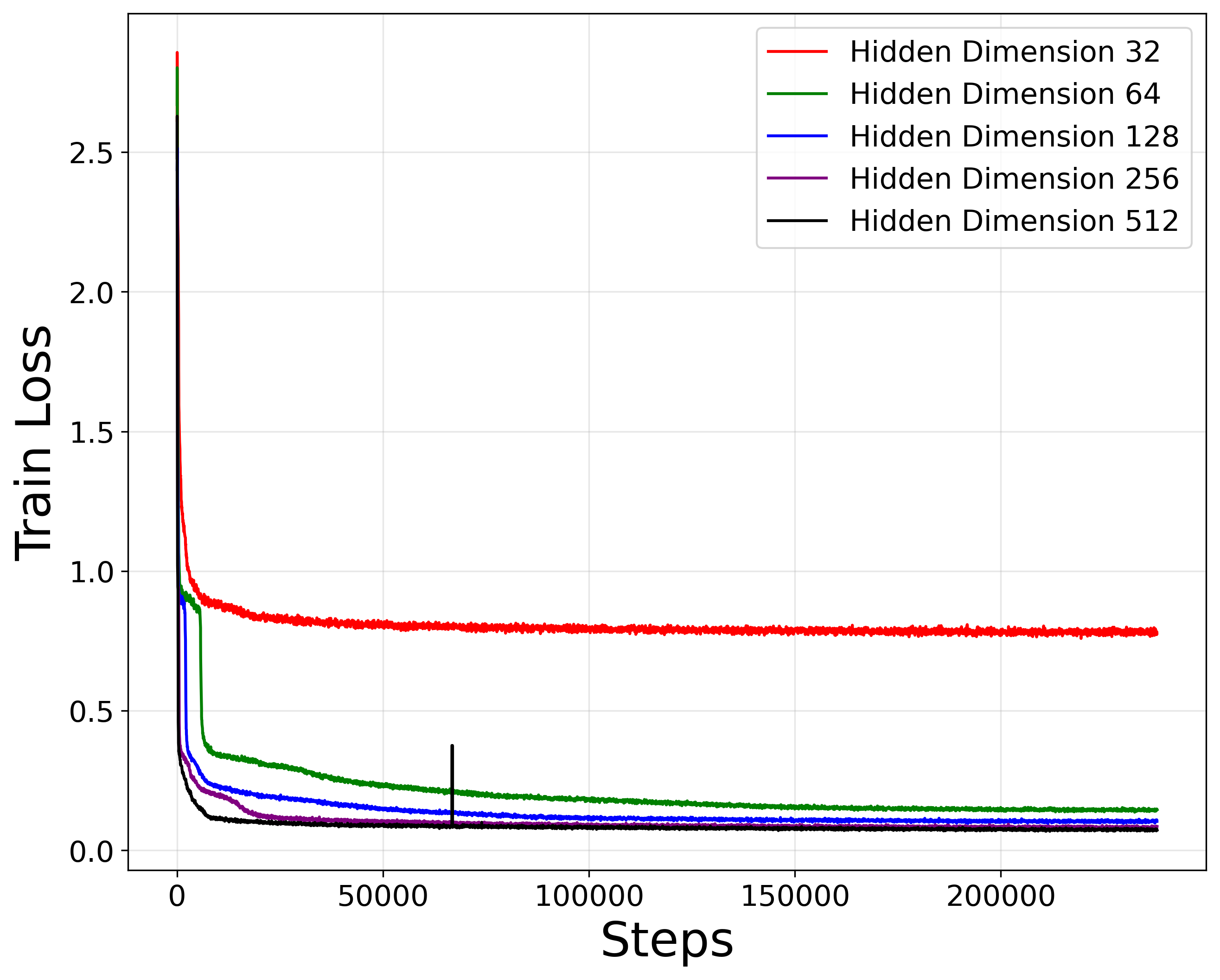}

  \caption{Ablation of the hidden dimension of the 2-layer model with $4$ attention heads. We predict that the model will fail when the maximum number of edges in a graph ($45$) exceeds the hidden dimension. The model with hidden dimension of $64$ and greater are able to reduce the loss and learn the task but $32$ is not.}\label{fig:hidden_dim_ablation}
\end{figure}
We test this prediction in Figure~\ref{fig:hidden_dim_ablation} with the same experimental setting described in Section~\ref{setting}. 2-layer models with $4$ attention heads and hidden dimension of $64$ and greater are able to reduce the loss and learn the task. However, the 2-layer model with a hidden dimension of 32 fails. We also point out that the greater the hidden dimension, the more quickly it can reduce the loss.

\begin{figure}[h]
 \centering
\includegraphics[width=0.45\textwidth]{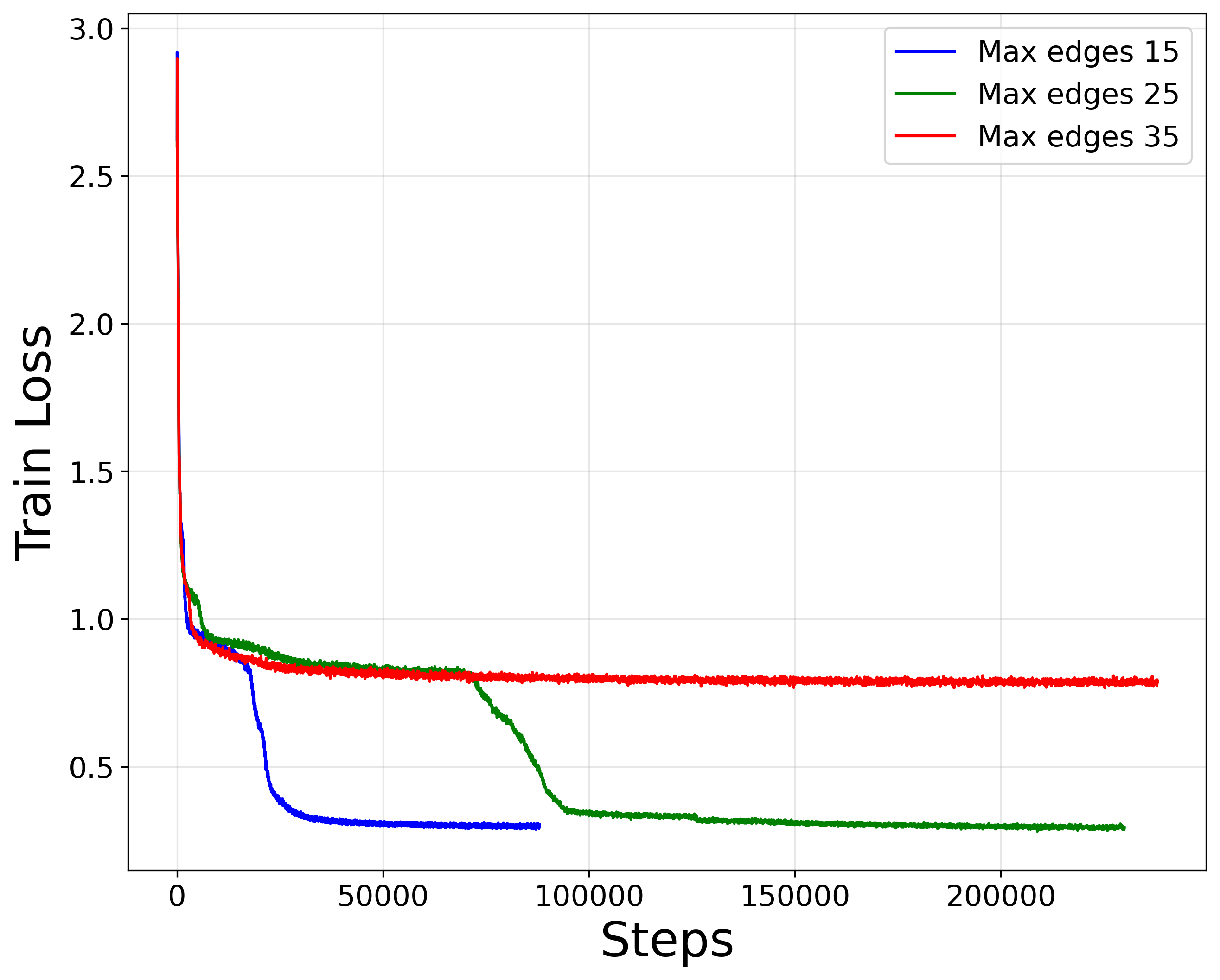}

  \caption{Ablation of the maximum number of edges in graphs in the training set. When we bound the number of edges to be less than the hidden dimension, the model is able to reduce the loss.}\label{fig:edge_ablation}
\end{figure}

A natural follow up experiment is to ablate the maximum number of edges in graphs in the training set. We present these results in Figure~\ref{fig:edge_ablation}. The 2-layer model with hidden dimension of $32$ is able to reduce the loss if we bound the maximum number of edges (e.g., $15$ and $25$) but at $35$ edges the model is still unable to learn. 

\subsection{Attention Heads}\label{app:attention}
We present additional results to complement the results in Section~\ref{activations}. We find $h_{current}$ and $h_{target}$ by manually expecting attention maps of each head in each model. Since the model distributes activation according to the degree of a node (which varies across samples), we normalize attention activations by dividing by the maximum value in a given sample. In Table~\ref{fig:head_values}, we present the activations of $h_{current}$ and $h_{target}$ averaged over the test set.
\begin{table}[h]
\begin{tabular}{|c|c|c|c|}
\hline
& Head 2 (/2) & Head 4 (/4) & Head 5 (/8) \\ \hline
Current&$0.826$&$0.75$&$0.779$\\ \hline
Other&$0.284$&$0.029$&$0.026$\\ \hline
Ratio &$2.91$&$25.86$&$29.96$ \\ \hline
& Head 1 (/2) & Head 1 (/4) & Head 1 (/8) \\ \hline
Target&$0.865$&$0.868$&$0.561$\\ \hline
Other&$0.45$&$0.036$&$0.007$\\ \hline
Ratio &$1.92$&$24.11$&$80.14$ \\ \hline
\end{tabular}
\caption{Table showing that each model learns heads which attend to current node/target node edges}
\label{fig:head_values}
\end{table}

\subsection{\oursfull}\label{app:sln}
\begin{figure}[H]
\centering
\includegraphics[width=0.5\textwidth]{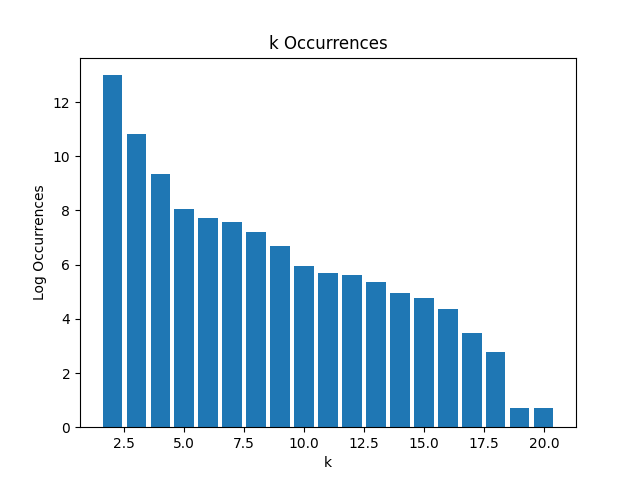}
\vspace{-.5cm}
\caption{The log of $k$ counts needed for \ours\ to successfully find the shortest path over the test set. For most graphs, using only the second smallest $k=1$ (i.e., Fiedler Vector) is sufficient.}\label{fig:k_counts}
\end{figure}
We implement and run \ours\ on the test set and are able to achieve a final accuracy of $99.32\%$. We found that for most graphs (roughly $80\%$) using only the second smallest eigenvalue (i.e., the Fiedler vector from Section~\ref{preliminaries}) for edge embeddings is sufficient. However, for other graphs we needed to increase the number $k$ of non-zero eigenvalues we consider for edge embeddings. We report the log of counts for each $1 \le k \le |E|$ in Figure~\ref{fig:k_counts}. 

\end{document}